# Vision System of Curling Robots: Thrower and Skip

Seongwook Yoon, Gayoung Kim, Myungpyo Hong, and Sanghoon Sull, *Senior Member, IEEE*

## I. INTRODUCTION

Intelligent robot systems are becoming increasingly popular with the rapid development of hardware and software technologies as well as artificial intelligence (AI). As one of recent successful achievements, AlphaGo defeated one of the best human go players, provoking concern that AI robots would be more powerful and smarter than human. However, AlphaGo just corresponds to the human brain without other abilities such as visual recognition and picking up go stones, for example. Therefore, in this article, we present our camera vision system of the AI curling robots designed to play curling game with human players.

Curling is one of the most popular winter sports game requiring complex strategy and physical abilities, thus attracting many researchers in the areas of artificial intelligence and robotics. There have been several researches on curling. Physics of curling games was analysed [1, 2, 3] and curling game simulation software was developed [4, 5]. Recently, there have been more focus on developing optimal strategies [6, 7, 8, 9]. The primitive curling robot consisting of a simple stone delivery mechanism and a simulator was developed [10]. Algorithms for tracking stones were also proposed [11, 12].

Our goal is to build a robot curling team corresponding to a human curling team without sweepers in the first version. In case of human players, a skip as the captain of a team recognizes stones, determines optimal strategy and indicates where a thrower on the opposite side should throw stone. Then, the thrower slides and releases stones before the hogline. Thus, as shown in Figure 1, a robot curling team consists of two identical robots which can play as both a skip and a thrower.

An AI curling robot should have three major abilities of recognizing curling game, thinking strategy and controlling robot behaviour. The first ability of a curling robot may be the recognition of curling game. Other two abilities require the result of recognition so the performance of the other two or even that of the whole robot would be bounded by that of recognition.

Thus, one of the most important major tasks of our curling robot is the recognition of curling game. The skip robot should localize stones around the house and track the stone thrown by the thrower robot. The thrower robot also needs to estimate its pose with respect to the sheet coordinate. These tasks should be able to be performed well without the external sensors attached to the sheet or the ceiling. Other types of internal sensors such as a gyroscope and a magnetic dip can be used to obtain relative angle changes for control. In this article, we propose a real-time camera vision system for AI curling robot. Our camera vision system involves three kinds of RGB camera with different lenses. The near-view camera with large field-of-view (FOV) observes the area around house to estimate the robot's pose using the curling pattern drawings such as circle and lines. Also, using the same camera, the robot estimates stones' position based on the circular boundary of their handle part. The far-view camera with smaller FOV magnifies the area farther than the hogline to track distant thrown stones. Those two cameras are mounted on the robot's head so that the robot should unfold its arm to recognize such information. The low-view camera with the largest FOV mounted on robot's body also estimates

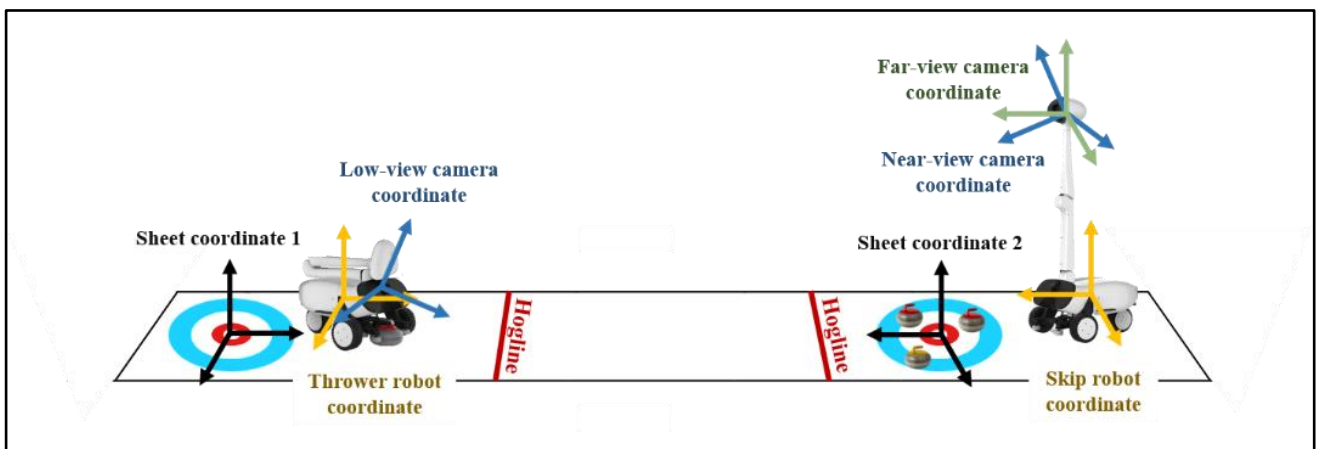

Figure 1: Our vision system mounted on the skip and thrower robots. The transformation between the camera coordinate and the sheet coordinate is determined by pose estimation using drawings on the curling sheet. The positions of stones are estimated with respect to the sheet coordinate system.

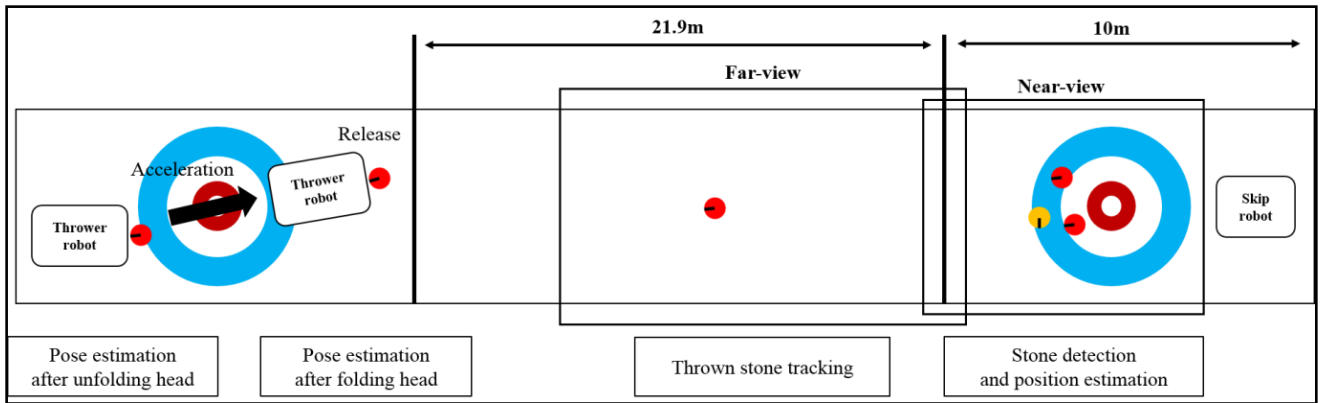

Figure 2: The four tasks to be performed by the vision system of curling robots.

the robot's pose while the robot folds its arm to run and release stone.

This article is organized as follows. In Section 2, we explain the configuration of the vision system. In Section 3, we describe our proposed vision system. In Section 4, we present the experimental results. Section 5 concludes the article.

## II. SYSTEM CONFIGURATION

Initially, the design requirements of AI curling robots were given as follows: i) The number of curling robots should be equal or less than that of the opponent team. ii) Curling robots adhere to the curling game rules. iii) Curling robots only operates during their turn. iv) Additional external sensors such as cameras mounted on the ceil, laser tracker and GPS are not used.

In the whole process of a turn of robot curling team, the vision system should perform the four tasks as shown in Figure 2. First, the skip robot should detect all stones nearer than hogline and estimate their positions with respect the sheet coordinate. Based on the estimated positions, an optimal strategy whose output is the optimal angle, speed and curl of the throwing stone is generated and transmitted to the thrower robot. At the same time, the thrower robot unfolds its arm to estimate the accurate angle between the sheet and itself. After horizontally rotating its body by internal sensors and folding its arm, the thrower robot starts to run and estimate the distance to the hogline by the low-view camera mounted on its body. Since the robot runs straight, the desired release point can be determined with only the distance. At the moment of release, a signal is transmitted to the skip robot. Then, the skip robot observes the thrown stone by the cameras of its head and tracks the trajectory of the thrown stone.

In order to perform the tasks described above, we use monocular RGB cameras rather than the depth sensors such as stereo cameras or LIDAR not only because monocular RGB cameras are cheaper but also because the accuracy of the estimated position is expected to be similar as long as the stone is constrained to be on the sheet plane. Such detection requires visual features of the stone such as the stone's center. Also, the pose with respect to the sheet coordinate can be estimated by the visual patterns such as hogline, sidelines and concentric rings in the house.

However, as the robot observes stones with the perspective view, the error of the stone position driven by the error of detecting stone's center fast increases with the distance. In order to solve this problem, we use the far-view camera covering the area farther than the hogline, as shown in Figure 3.

The near-view and far-view cameras are mounted at the height of 2m as shown in Figure 3(a). Since the thrower robot cannot stably slide on the ice sheet with its arm unfolded, it folds its arm as shown in Figure 3(c). The pose estimation is also needed during the acceleration and release and, thus, a single monocular camera called as low-view camera is mounted at the robot body of 0.5m height.

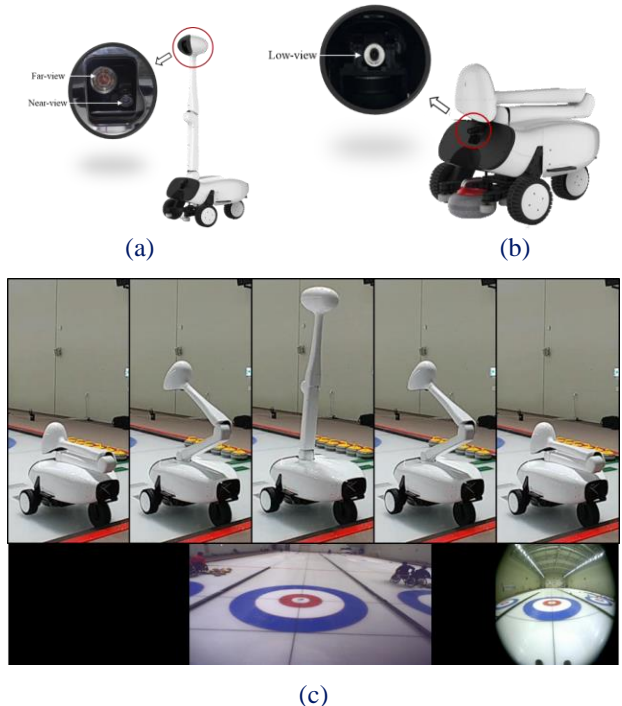

Figure 3: Three cameras on each robot. (a) The far-view and near-view cameras operating only when the robot unfolds its arm. (b) The low-view camera operating when the robot folds its arm to accelerate to throw a stone. (c) Pose estimation before it acceleration of the thrower robot through the process of unfolding and folding its arm. The images below are examples of frames captured by the near-view camera after unfolding its arm, and the low-level camera, respectively.

## III. PROPOSED VISION SYSTEM

In this section, we describe the details of our vision system. First, we explain the pose estimation using the near-view camera including curling pattern detection. Second, we explain another pose estimation using the low-view camera. Third, we explain detection and position estimation of stones. Last, we present how to track the thrown stone.

### A. Pose estimation using the near-view camera

Pose estimation is needed for both the thrower robot and skip robot. The pose estimation computes extrinsic parameters for the transformation between the camera coordinate and the sheet coordinate. In case of the thrower robot, we estimate the thrower robot's position (translation) and direction (rotation) with respect to the sheet coordinate. This gives an initial yaw angle between the thrower robot and the curling sheet so that the robot can rotate and run toward the desired direction. However, the position estimated by the vision system is not used because the robot cannot move freely enough to get to the right position. The allowable error of the angle estimate is less than 0.2 degrees so we need to develop a novel and precise method.

Pose is estimated by the well-known Perspective-n-Point solution [13] of which each pairs of points are 2 coordinates of the image plane and 3 coordinates of the sheet coordinate. The problem is how to choose the good feature points. In this article, we utilize the patterns on the curling sheet. First, we observe that the pattern involves a hogline (near one), two sidelines (left and right ones) and a circle (the outermost one of concentric rings in the house) at each side of a curling sheet. It is noted that the sidelines could be very thin or even omitted for some curling sheets. While there are stringent rules about the hogline and the circle, about 6.4% deviation is allowed for the sidelines. Thus, the sidelines are not used directly for pose estimation in this article, but they give us a clue on the forward direction of the sheet.

Our idea is to find the feature points which can be determined by geometric characteristics of the planar projection. As we deal with the lines and circles, the four points can be reliably found as illustrated in Figure 4.

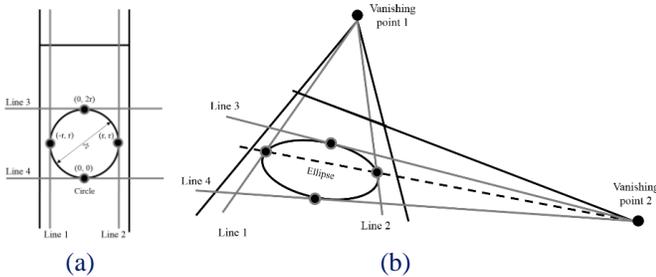

Figure 4: Illustration of picking the four points from the projection of curling sheet. It is noted that (b) is exaggeratedly drawn for illustration purpose. The lines 1 and 2 in (a) are projected to the lines 1 and 2 as in (b). The lines 1 and 2 in (b) can be obtained as the tangent lines from the vanishing point 1 to the ellipse, and we draw a virtual t-line (dotted line) by connecting two points of contact. Similarly, the lines 3 and 4 in (b) can be calculated as the tangents lines from the vanishing point 2 to the ellipse. As a result, we can obtain the four points on the image with the corresponding sheet coordinates because we can assign coordinates to the four points as shown in (a).

Thus, we can accurately estimate the pose if the sidelines, hogline and ellipse are detected precisely from the image. However, the reliable detection of the correct lines and circles is not an easy problem. In this article, we solve this problem by choosing the best combination minimizing the reprojection error obtained by the Perspective-n-Point algorithm after detecting candidate lines from the image.

We used several conventional ways of detecting lines and contours of an ellipse projected from a circle. Our process to extract candidates includes the correction of lens distortion of an image as in [14], detection of lines using Hough Line Transform, classification of lines, color segmentation, detection of contours as in [15], and the selection of ellipse contours from the contours.

Most lines can be easily detected by the voting algorithm based on Hough Line Transform [16]. Although general classification of detected lines is non-trivial, we simply use the geometric features such as slope and intercept since we know the approximate values of pose, enabling us to classify lines into four classes of hogline, left sideline, right sideline and none. Color segmentation basically divides the house into 3 colored regions since the house usually consists of white ice, outer ring and inner ring of different colors. The region to which the color segmentation is applied is determined by the hogline, the left sideline and the right sideline. The area bounded by the most probable candidate three lines is colored by only three colors and thus can be segmented easily with the simple k-means algorithmic color segmentation method. We can find the contour of the outer ring from the segmented regions and fit an ellipse to the contour in order to calculate the four points.

Finally, the best candidate $d$ is selected from the extracted candidates by finding the set minimizing the following error:

$$d = \underset{\substack{(h,l,r) \in H \times L \times R \\ h \neq l \neq r}}{\arg\min} \; E_{reprojection}(P_4(h,l,r)), \quad (0.1)$$

where $H$ is a set of candidates of hoglines, $L$ is a set of candidates of left sidelines, $R$ is a set of candidates of right sidelines, $P_4$ is a function of computing 4 points as in Figure 4, and $E_{reprojection}$ represents a function of the 4 points calculating the reprojection error by using the pose estimated from the Perspective-n-Point algorithm. The reprojection error is the difference between the image points obtained as Figure 4(b) and the image points transformed from the sheet coordinates so that it can be referred to as a criterion of the points' feasibility. In practice, we add a constraint re-verifying whether the pose estimated by Perspective-n-Point is included in the appropriate range.

### B. Pose estimation using the low-view camera

After the pose estimation of the near-view camera, the thrower robot folds its arm and starts to run. The goal of the thrower robot is to release stone at the position, angle and speed indicated by the strategy module. To achieve this, other internal sensors were tried but the position cannot be precisely measured by those sensors. Thus, we use our vision system to estimate the position of the thrower robot as the process of the pose estimation of the low-view camera mounted on the body part of the robot.

To estimate the position, we utilize the hoglines and two sidelines. Thus, we first detect line candidates and compute their geometry with respect to the camera coordinate using the pre-measured tilt and height of the low-view camera. To detect the correct hogline and sidelines, we can utilize the value of sheet width obtained by the pose estimation using the near-view camera. Thus, we detect three lines corresponding to the shape of letter H by minimizing an error defining the feasibility of the shape. The error consists of three error terms; two terms are shown as the red segments in Figure 5 and third one is the difference between the distance of two interceptions and the value of the sheet width. Then, we simply calculate the rotation and translation from the two interceptions which minimize the error. The estimated distance to the hogline is accurate enough for the robot to release the stone at the indicated position because the robot also run straight enough.

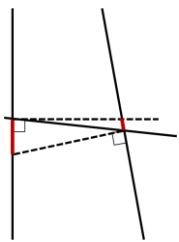

Figure 5: The errors used to detect hogline and sidelines.

### C. Stone detection and position estimation using the near-view camera

Detection of stones and estimation of their positions with respect to the sheet coordinate are performed by the skip robot. Like the pose estimation, the near-view camera observes the area nearer than the hogline and detects the stones nearer than the hogline. Actually, those stones are the entire stones to be detected in the sheet until a stone is thrown.

The detection and estimation are not easy due to the occlusion between stones. The stones appearing in most frame images are occluded if other stones are close as shown in Figure 6. Therefore, we focus on their upper parts mostly showing their handles since the handles are usually visible without occlusion.

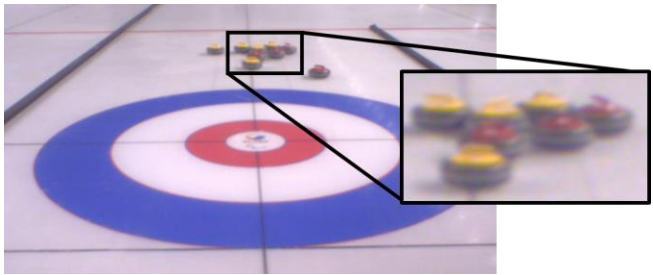

Figure 6: Occlusion between stones.

The handle, precisely the circular boundary of the handle, is not entirely occluded so that we can always identify the pixels corresponding to the center. Furthermore, the boundary exhibits distinct edges in RGB images because the handle has chromatic colors. Our goal is simply to detect the center of the circular boundary and calculate their position with its known height and the robot's pose estimated with the near-view camera.

The first step of our method is to detect a coarse bounding box of each group of stones which are closely located. This step is first performed prior to any other process and does not require accurate segmentation of each stone in the bounding box.

The second step is the pose estimation of the skip robot. This step performs the same task as the pose estimation of the thrower robot except that there would be stones in the sheet. We simply exclude all edges around or inside the detected bounding box from the voting step for extracting candidates of patterns. Because the stones look relatively smaller than the curling sheet pattern, detecting the sheet pattern is still possible.

The third step is to estimate the number of stones and their positions in each bounding box. We use the voting algorithm based on Hough Circle Transform. Though there are many ways of detecting ellipses such as [17, 18], the ellipses according to the boundary is the projection of a circle of a fixed height and size. All the edges in each bounding box are projected to the sheet plane of the stone's height and are voted for the parameters of circles. Since we know the actual radius of the circle, the parameter is only the coordinates of its center with respect to the sheet. Thus, the peak point whose voting score is the maximum in the area corresponding to the bounding box is the most probable point of the center of a stone. If the voting score of the point is larger than a predetermined threshold, it is selected as the center of a stone. Then, all the points nearer than the diameter of the stone with the center are removed. These two sub-steps of selection and removal are repeated until there are no points exceeding the threshold.

### D. Thrown stone tracking using the near-view and far-view cameras

The skip robot is also responsible for tracking the thrown stone where the trajectory is used to adjust the strategy for the next stone to throw by analyzing ice condition. While the detection of the thrown stone is nearly equal to the detection of stationary stones, we utilize both the near-view and far-view cameras and also check the collision of the thrown stone. Checking whether the thrown stone stops after collision with the other stone or not can verify whether the position on which the thrown stone stops is useful to analyze the ice condition.

Our tracking algorithm for the thrown stone is based on Kalman filter. The tracking of multiple stones for collision checking is based on the identification of each stone at each frame and the distance between stones.

The far-view camera is used to tracking of the thrown stone farther than the hogline and the near-view camera is used after the thrown stone passes the hogline. The reason why two cameras are switched at the moment which is the first chance of collision is because we need to observe all stones around the thrown stone to check collision.

There are two steps for verifying whether the thrown stone is collided to stationary stones. The first step is to check whether the thrown stone moves around by calculating the distance between the thrown stone's position tracked every frame and the stationary stone's position estimated before. Next step is to newly track the stationary stones close enough to the thrown stone. If the newly tracked stone is determined to be moving, the thrown stone collided to the stone. In addition to the collision, our algorithm checks whether the thrown stone gets out of the sheet.

As the result of this tracking, we have the trajectory of the thrown stone before it collides to other stones and its final stationary position. We also calculate the time taken for the thrown stone to pass the near hogline from the far hogline, which is called a hog-to-hog timing. The value is commonly used by the human players and robust enough to be applied directly to the adjustment of strategy.

## IV. EXPERIMENTAL RESULTS

In this section, we describe the implementation details and experimental results of the proposed system.

### A. Implementation details

Our curling robot has four major modules of strategy, control, communication and vision. In this subsection, we describe details about the vision module and its related parts.

The vision module in each robot consists of three cameras and one board for real-time image processing. We also used the image acquisition system UI-3250CP-C-HQ Rev.2. with the same values of their parameters excluding the gains for colour calibration. The main configuration difference is lenses used for each camera. The lens of the near-view camera should have at least 100 degrees of horizontal FOV, considering the width of the sheet and the height of the camera. Since the FOV of the far-view camera should be determined considering the trade-off between the pixel resolution and the range of skip robot's direction, we selected the adjustable zoom lens for the far-view camera such as M6Z1212-3S. For the low-view camera, we used the lens FV0420 with the larger FOV than other cameras. The image processing board used is the Jetson TX2 Developer Kit because it provides OpenCV4Tegra which accelerates general OpenCV codes with its Tegra chip.

Other modules are built in different devices connected by the communication module using the simple wired/wireless socket communication. The control module is built on two devices. A single board computer controls gripper which grips a stone or release the stone and the arm which folds or unfolds its robot head. An embedded device, myRio, controls the driving and traction. The strategy module is built on an external computer with high computational power. The communication module has several access points and includes the communication parts of each devices.

### B. Experiments

In this subsection, we describe the experimental results of our algorithms.

Figure 7 shows several examples of the detection results of the curling pattern for pose estimation using the near-view cameras of both thrower robot and skip robot. The lines and circle are detected correctly, and the four points calculated by our algorithm also corresponds to the true points in the image. The detection is robust to the robot's rotation, slight occlusion and the surroundings of the sheet. Especially, it is robust to the curling stones on the sheet which even occlude the pattern or the feature points.

To evaluate the actual performance of the pose estimation of the near-view camera, the yaw angle between the sheet coordinate and the thrower robot was also measured by another internal sensor, DSP 3000 gyrosensor. We compared the increment of our angle estimate with the increment of the measurements of the sensor while robot the rotates. We denote our angle estimate as $E$ in degrees, the gyrosensor measurement as $M$ in degrees and their increments as $\Delta E$ and $\Delta M$, respectively. Figure 8(a) shows that the difference, $\Delta E - \Delta M$, increases with the estimate $E$ around a bias. So, we needed to initially align the robot with the angle around the bias. We assumed that the bias is caused by the difference between the angle estimate and the true angle. So, we aligned the robot toward the center line in Figure 8(b) based on the estimated value of $E=0$ and investigated how the robot runs along the center line when it is ordered to do. Then, we calculated the bias by the angle which can be calculated by the distance from the center line when it passes the hogline.

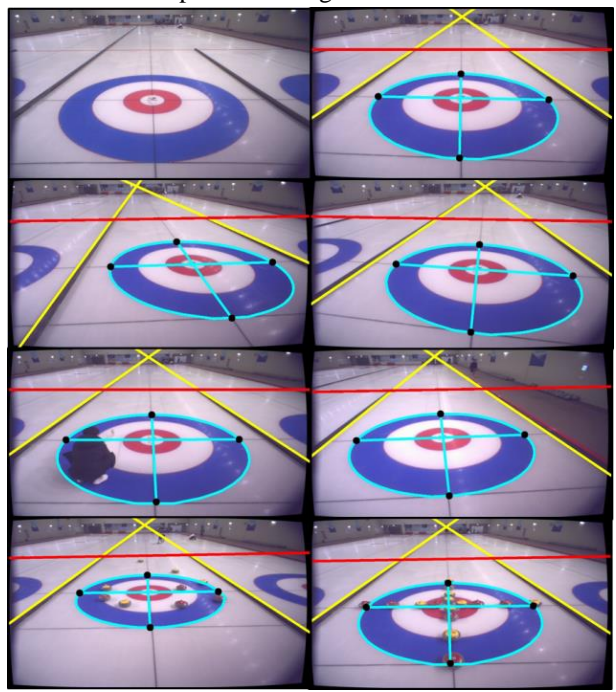

Figure 7: The detection results of the curling pattern is shown. The top-left image is an input image of the curling sheet and the detected curling pattern is overridden as thick lines and the four detected feature points are represented as black points.

The accuracy of the pose estimation using the low-view camera was manually checked because there was no accurate way to measure the true pose of the running robot. The system estimates the distance from the hogline in real-time and releases the stone considering the velocity and latency of the estimation. We repeated the stone release processes and confirmed that the distance between the release point and the hogline was always around 50 centimeters as desired. Figure 9 shows the result as the robot moves along for several frames.

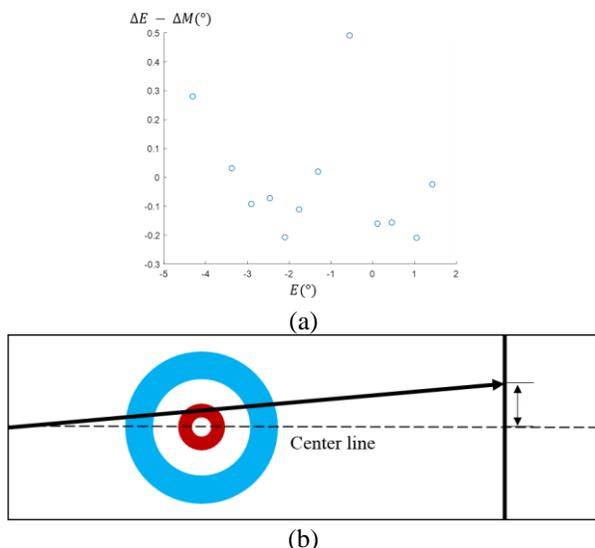

(a)

(b)

Figure 8: (a) The difference between the estimates using the camera and the measurement using DSP 3000 varies along the angle estimate $E$. (b) The robot runs along the arrowed line if the bias exists.

The accuracy of the detection and position estimation of the stone was difficult to be measured because the ground truth of the stone position is hard to obtain without overlaying a dense grid on the sheet. Instead, we located stones in position of which coordinates can be measured directly by the cross pattern as shown in first example of Figure 10. According to several repeated such experiments, the error of the position estimation of stones in the house was smaller than 5 centimeters. The other examples in Figure 10 show that the algorithm performs well on many of probable scenes including the occlusion between the stones.

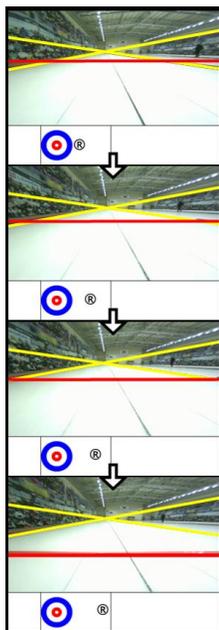

Figure 9: The detection result of the three lines using the low-view camera is shown in this figure. Two sidelines are denoted with yellow line and hogline is denoted with red line. The estimated position of the throwing robot is represented as R in the sheet.

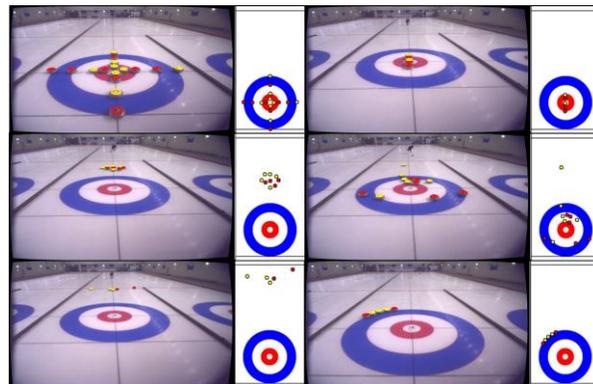

Figure 10: The results of stone detection and position estimation are shown in both image and sheet. In image, the detected boundary and center of stone's handle part is colored with each stone's color. In sheet, colored circles are visualized in the estimated positions.

The tracking accuracy of the thrown stone was more difficult to be measured than that of estimating the position of stationary stones. Since the position estimation of our tracking algorithm is identical to estimating the position of stationary stones, we located a stone in several positions and measured its coordinates manually. We investigated the tendency of the accuracy with respect to the distance from the skip robot. As expected, there is a rapid change of error values around the hogline because the camera change occurs when the thrown stone passes the hogline. The error of the nearest stone was 5 centimeters as we mentioned above and the error of the position estimation increases linearly in the area observed by each camera. The error of the nearest area of the far-view camera (about 10 centimeters), was smaller than the error in the farthest area of the near-view camera (about 20 centimeter). An example of trajectory is shown in Figure 11.

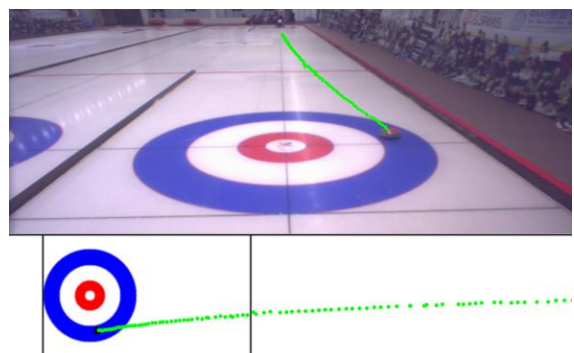

Figure 11: Trajectory of a thrown stone.

Throughout the development of our robots, our skip and thrower robots had played several games with two competitive amateur curling players and a national wheelchair curling player team. The robot team played competitive games, but we lost several games especially due to the varying ice conditions that we did not expect. For example, humidity increased due to much more people than allowed in the curling rink. However, the vision algorithms performed well in the games even though there were many peoples around the sheet. Figure 12 shows 6 scene of the curling games with human players.

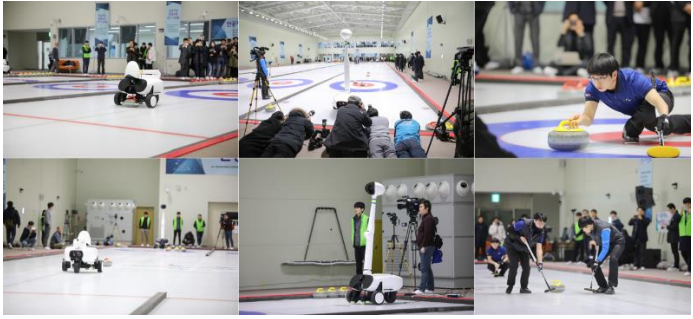

Figure 12: A game of our robot with human players in March, 2018.

## V. CONCLUSION

We developed a camera vision system of AI curling robots recognizing a curling game. It is practically well-designed so that the curling robots can competitively play with human amateur player teams and they even won a couple of games. The remarkable points of our system are that we use only three RGB cameras mounted on the robots without ceiling cameras or other external sensors and that our system accurate operates in real-time.

Our consortium plans to build sweeping robots this year. From the vision system point of view, we need to devise s method of estimating the position of the sweeping robots. We plan to increase the accuracy of the position of the thrown stone when it is farther from the skip robot. We also plan to make vision system more robust with respect to the possible occlusion caused by the sweeping robots.

**Acknowledgments** This work was supported by Institute for Information & Communications Technology Promotion (IITP) grant funded by the Korea government (MSIT) (No. 2017-0-00521).

# References

[1] Maeno, Norikazu. "Curl mechanism of a curling stone on ice pebbles." *Bulletin of Glaciological Research* 28 (2010): 1-6.

[2] Maeno, Norikazu. "Dynamics and curl ratio of a curling stone." *Sports Engineering* 17.1 (2014): 33-41.

[3] Nyberg, Harald, et al. "The asymmetrical friction mechanism that puts the curl in the curling stone." *Wear* 301.1-2 (2013): 583-589.

[4] Heo, Myeong-Hyeon, and Dongho Kim. "The development of a curling simulation for performance improvement based on a physics engine." *Procedia Engineering* 60 (2013): 385-390.

[5] Ito, Takeshi, and Yuuma Kitasei. "Proposal and implementation of" digital curling"." *Computational Intelligence and Games (CIG), 2015 IEEE Conference on*. IEEE, 2015.

[6] Yamamoto, Masahito, Shu Kato, and Hiroyuki Iizuka. "Digital curling strategy based on game tree search." *Computational Intelligence and Games (CIG), 2015 IEEE Conference on*. IEEE, 2015.

[7] Ohto, Katsuki, and Tetsuro Tanaka. "A Curling Agent Based on the Monte-Carlo Tree Search Considering the Similarity of the Best Action Among Similar States." *Advances in Computer Games*. Springer, Cham, 2017.

[8] Won, Dong-Ok, et al. "Curly: An AI-based Curling Robot Successfully Competing in the Olympic Discipline of Curling." *IJCAI*. 2018.

[9] Lee, Kyowoon, et al. "Deep Reinforcement Learning in Continuous Action Spaces: a Case Study in the Game of Simulated Curling." *International Conference on Machine Learning*. 2018.

[10] Kawamura, Takashi, et al. "A study on the curling robot will match with human result of one end game with one human." *Computational Intelligence and Games (CIG), 2015 IEEE Conference on*. IEEE, 2015.

[11] Kim, Junghu, et al. "Curling Stone Tracking by an Algorithm Using Appearance and Colour Features." *Proceedings of the World Congress on Electrical Engineering and Computer Systems and Science*. Vol. 334. 2015.

[12] Takahashi, Masaki, et al. "Visualization of Stone Trajectories in Live Curling Broadcasts using Online Machine Learning." *Proceedings of the 2017 ACM on Multimedia Conference*. ACM, 2017.

[13] Lepetit, Vincent, Francesc Moreno-Noguer, and Pascal Fua. "Epnp: An accurate o (n) solution to the pnp problem." *International journal of computer vision* 81.2 (2009): 155.

[14] Zhang, Zhengyou. "A flexible new technique for camera calibration." *IEEE Transactions on pattern analysis and machine intelligence* 22.11 (2000): 1330-1334.

[15] Suzuki, Satoshi. "Topological structural analysis of digitized binary images by border following." *Computer vision, graphics, and image processing* 30.1 (1985): 32-46.

[16] Duda, Richard O., and Peter E. Hart. "Use of the Hough transformation to detect lines and curves in pictures." *Communications of the ACM* 15.1 (1972): 11-15.

[17] McLaughlin, Robert A. "Randomized Hough Transform: Improved ellipse detection with comparison1." *Pattern Recognition Letters* 19.3-4 (1998): 299-305.

[18] Xie, Yonghong, and Qiang Ji. "A new efficient ellipse detection method." *Pattern Recognition, 2002. Proceedings. 16th International Conference on*. Vol. 2. IEEE, 2002.

**Seongwook Yoon**, School of Electrical Engineering, Korea University, Seoul, Korea. E-mail: swyoon@mpeg.korea.ac.kr.

**Gayoung Kim**, School of Electrical Engineering, Korea University, Seoul, Korea. E-mail: gykim@mpeg.korea.ac.kr.

**Myungpyo Hong**, School of Electrical Engineering, Korea University, Seoul, Korea. E-mail: mphong@mpeg.korea.ac.kr.

**Sanghoon Sull**, School of Electrical Engineering, Korea University, Seoul, Korea. E-mail: sull@korea.ac.kr.